\documentclass[11pt,a4paper]{article}
\usepackage{times,latexsym}
\usepackage{url}
\usepackage{amsmath}
\usepackage{amsthm}
\usepackage{amsfonts}
\usepackage{amssymb}
\usepackage{mathtools}
\usepackage[T1]{fontenc}

\usepackage{tikz}
\usetikzlibrary{bayesnet}
\usetikzlibrary{arrows}
\usetikzlibrary{decorations.markings}
\usetikzlibrary{snakes}

\usepackage{graphicx}
\usepackage{booktabs}
\usepackage{tabularx}
\usepackage{tabulary}
\usepackage{colortbl}
\usepackage{xspace}
\usepackage{xcolor}
\newcolumntype{Y}{>{\centering\arraybackslash}X}
  \newcolumntype{P}{>{\raggedleft\arraybackslash}X}
\usepackage{multirow}
\usepackage{stfloats}

\usepackage{algorithm}
\usepackage[noend]{algorithmic}

\usepackage{enumitem}

\usepackage[acceptedWithA]{tacl2018v2}

\usepackage[noabbrev,capitalise]{cleveref}

\usepackage{xspace,mfirstuc,tabulary}

\newif\iftaclinstructions
\taclinstructionsfalse 
\iftaclinstructions
\renewcommand{\confidential}{}
\renewcommand{\anonsubtext}{(No author info supplied here, for consistency with
TACL-submission anonymization requirements)}
\newcommand{\instr}
\fi

\iftaclpubformat 

\else

\fi

\newcommand{\calT}{{\cal T}}
\newcommand{\calD}{{\cal D}}
\newcommand{\calS}{{\cal S}}

\newcommand{\vphi}{{\boldsymbol \varphi}}
\newcommand{\vtheta}{{\boldsymbol \vartheta}}

\newcommand{\xx}{\mathbf{x}}

\newcommand{\zz}{\mathbf{z}}

\newcommand{\bb}{\mathbf{b}}

\DeclarePairedDelimiterX{\infdivx}[2]{(}{)}{%
  #1\;\delimsize\|\;#2%
}

\newcommand{\skilled}{\textbf{\sc {Skilled}}\xspace}
\newcommand{\shared}{\textbf{\sc {Shared}}\xspace}
\newcommand{\private}{\textbf{\sc {Private}}\xspace}
\newcommand{\expert}{\textbf{\sc {Expert}}\xspace}
\newcommand{\hyperformer}{\textbf{\sc {Hyper\-Former}}\xspace}

\usepackage{todonotes}
\makeatletter
\newcommand*\iftodonotes{\if@todonotes@disabled\expandafter\@secondoftwo\else\expandafter\@firstoftwo\fi}
\makeatother

\definecolor{edolime}{rgb}{0.9,1,0.3}

\definecolor{Gray}{gray}{0.92}

\title{Combining Modular Skills in Multitask Learning}

\newcommand{\mila}{\mu}
\newcommand{\msr}{\sigma}
\newcommand{\mcgill}{\gamma}
\newcommand{\umontreal}{\rho}

\author{\bf Edoardo M. Ponti$^{\mila,\mcgill}$~\;~Alessandro Sordoni$^{\msr}$~\;~Yoshua Bengio$^{\mila,\umontreal}$~\;~Siva Reddy$^{\mila,\mcgill}$ \\
$^{\mila}$Mila -- Quebec AI Institute~\;~$^\mcgill$McGill University \\ $^\msr$Microsoft Research Montr\'eal~\;~$^\umontreal$Universit\'e de
Montr\'eal \\
$^{\mila}$\texttt {\{edoardo-maria.ponti,yoshua.bengio,siva.reddy\}@mila.quebec} \\
$^{\msr}$\texttt {alsordon@microsoft.com}
}

\date{}

\begin{document}
\maketitle

\begin{abstract}
A modular design encourages neural models to disentangle and recombine different facets of knowledge to generalise more systematically to new tasks.
In this work, we assume that each task is associated with a subset of latent discrete skills from a (potentially small) inventory. In turn, skills correspond to parameter-efficient (sparse / low-rank) model parameterisations. By jointly learning these and a task--skill allocation matrix, the network for each task is instantiated as the average of the parameters of active skills.
To favour non-trivial soft partitions of skills across tasks, we experiment with a series of inductive biases, such as an Indian Buffet Process prior and a two\-speed learning rate.
We evaluate our latent-skill model on two main settings: 1) multitask reinforcement learning for grounded instruction following on 8 levels of the BabyAI platform; and 2) few-shot adaptation of pre-trained text-to-text generative models on CrossFit, a benchmark comprising 160 NLP tasks. We find that the modular design of a network significantly increases sample-efficiency in reinforcement learning and few-shot generalisation in supervised learning, compared to baselines with fully shared, task-specific, or conditionally generated parameters where knowledge is entangled across tasks. In addition, we show how discrete skills help interpretability, as they yield an explicit hierarchy of tasks.
\end{abstract}

\section{Introduction}
Modularity endows neural models with an inductive bias towards systematic generalisation, by activating and updating their knowledge sparsely \citep{hupkes2020compositionality}. For instance, modules may compete to attend different parts of a structured input \citep{goyal2019recurrent,goyal2020object} or disentangle pre-determined skills, reusable and autonomous facets of knowledge, across multiple tasks to be later recombined in original ways for new tasks \citep{alet2018modular,ponti2021inductive,ansell-etal-2021-mad-g,kingetsu2021neural}. These two levels of modularity mirror two distinct levels of memory \citep{hill2020grounded,yogatama2021adaptive}---short-term for input-level knowledge and long-term for task-level knowledge---and ultimately reflect the integrated yet modular nature of the cognitive system in humans \citep{clune2013evolutionary}.

\begin{figure}[t]
    \centering
    \includegraphics[width=\columnwidth]{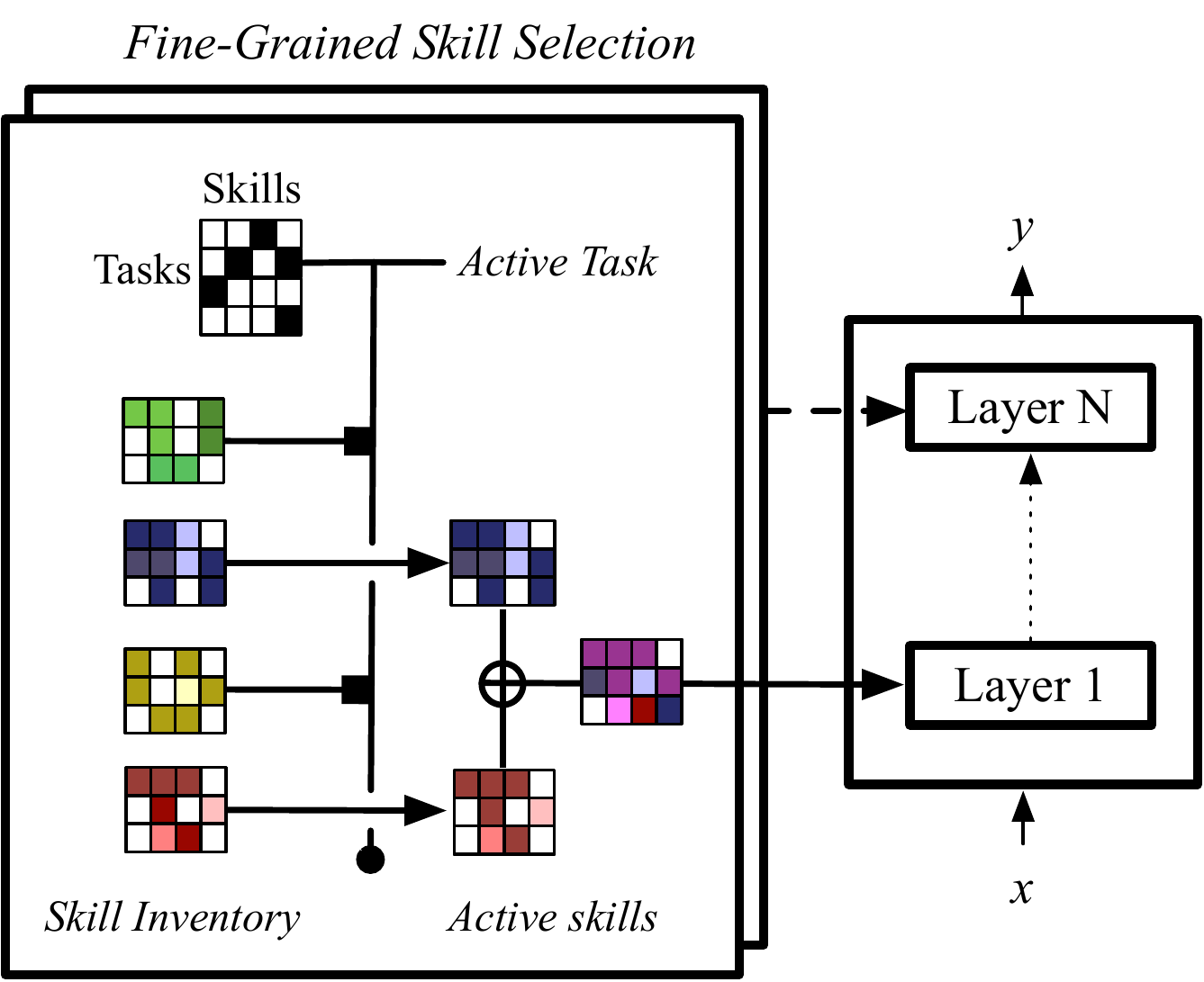}
    \caption{A diagram of our latent-skill model: 1) a row of the task--skill binary matrix is selected according to the active task; 2) the (sparse or low-rank) parameters corresponding to active skills from a layer-specific inventory are combined; 3) the resulting parameterisation is plugged into a neural network.}
    \label{fig:diag}
\end{figure}

In multitask learning, previous work focused on settings where the skills relevant for each task are known \textit{a priori} \citep{pfeiffer-etal-2020-mad,ponti2020parameter,ansell2021composable}, or settings where parameters or representations are an entangled mixture of shared and task-specific knowledge \citep{misra2016cross,ruder2019latent,mahabadi2021parameter}. The former method requires expert knowledge (possibly sub-optimal and limited to a few domains), whereas the latter leaves multitask models vulnerable to distribution shifts and hence hinders them from quickly adapting to new tasks. To remedy these shortcomings, in this work we address a setting where the skills needed for each task are modular but unknown and possibly finer-grained than those posited by experts. Thus, we propose a method to jointly learn, in an end-to-end fashion: 1) a task--skill allocation matrix, which indicates which subset of skills (from a fixed inventory) are active for which task; and 2) a corresponding set of parameter-efficient adaptations of the model, a subset of which are superimposed to a base model according to the active skills.

The proposed model is equivalent to performing a soft partition, represented by a binary matrix, of the set of skill-specific parameters \citep{orbanz2012lecture}. Thus, an Indian Buffet Process \citep{griffiths2005infinite,teh2007stick} can be posited as a prior over this matrix, to regularise it towards striking a balance in allocating different subsets of skills to each task. As an alternative inductive bias, we also explore using a higher learning rate for the task--skill matrix compared to the skill-specific parameters, as it promotes better allocations over general-purpose parameterisations.

We evaluate our model on both reinforcement and supervised learning. For the first set of experiments, we focus on BabyAI \citep{chevalier-boisvert2018babyai}, a platform consisting of a sequence of levels where agents must follow linguistic instructions in a simulated environment. Each level is procedurally generated to reflect an increasingly complex combination of skills (e.g., picking up objects or unlocking doors). We find that our modular network achieves higher sample efficiency---by requiring a significantly smaller number of episodes to reach a near-perfect success rate in all levels---compared to all baselines including fully shared and level-specific models based on a state-of-the-art architecture \citep{hui2020babyai}, as well as a model that has access to the ground-truth skills.

In addition, in the wake of the recent surge of interest in massively multitask few-shot NLP models \citep[\textit{inter alia}]{min2021metaicl,wei2021finetuned,aribandi2021ext5,sanh2022multitask,mahabadi2021parameter}, we also evaluate our latent-skill model on CrossFit \citep{ye-etal-2021-crossfit}. This benchmark recasts 160 NLP tasks (including QA, conditional text generation, classification, and other types such as regression) as text-to-text generation problems. We obtain superior performance in few-shot adaptation to held-out tasks compared to a series of competitive baselines that include HyperFormer, a state-of-the-art method for task-conditional parameter generation \citep{mahabadi2021parameter}. Thus, we demonstrate the ability of our model to successfully reuse and adjust the skills, which were previously acquired during multitask pre-training, on new tasks.

Moreover, we show that our method can be used in tandem with several parameter-efficient methods \citep{he2021towards} in order to make the increase in time and space complexity due to skill-specific parameters negligible. In particular, we explore sparse adaptation with Lottery-Ticket Sparse Fine-Tuning \citep[LT-SFT;][]{ansell2021composable} and low-rank adaptation with Low-Rank Adapters \citep[LoRA;][]{hu2021lora}. Finally, in addition to sample efficiency and systematic generalisation, we illustrate how our method also favours interpretability, as it discovers explicitly which pairs of tasks require common modules of knowledge. 

The code for our model is available at: \\ \url{github.com/McGill-NLP/polytropon}.

\section{A Latent-Skill Multitask Model} 

The goal of multitask learning in modelling a set of tasks $\calT = (\calT_1, \dots, \calT_{|\calT|})$ is two-fold: 1) increasing sample efficiency on each seen task by borrowing statistical strength from the others; and 2) attaining systematic generalisation, the ability to adapt robustly to new tasks, possibly based on a few target-domain examples. In particular, in supervised learning, each task $\calT_i$ is associated with a dataset $\calD_i \triangleq \{(\xx_{1}, y_{1}), \dots, (\xx_{n}, y_{n})\}$ and a loss function $\mathcal{L}(\hat{y}, y)$, where each $\xx$ is an input and each $y$ is a label. In reinforcement learning, each task is characterised by an initial state distribution $q(\xx_1)$, a transition distribution $q(\xx_{t+1} \mid \xx_t, a_t)$, and a loss function $\mathcal{L}(\xx_1, a_1, \dots, \xx_h, a_h) \rightarrow \mathbb{R}$,\footnote{This is the negative of the reward: $\mathcal{L}(\cdot) = - \mathcal{R}(\cdot)$.} where $\xx$ is a state, $a$ is an action, and $h$ is the temporal horizon of each episode. Thus, each task defines a Markov Decision Process (MDP).

Previous work sought to counter the limitations of fully sharing the parameters of a model across all tasks \citep{pmlr-v97-stickland19a}, which may exhaust model capacity and lead to interference among the task-specific gradients \citep{wang2021gradient}.
Instead, parameters can be softly shared across tasks by composing task-specific adapters
~\citep{pfeiffer2020adapterfusion} or generating parameters via task-conditioned hyper-networks \citep{ponti2020parameter,mahabadi2021parameter,ansell-etal-2021-mad-g}. The first method leads to an explosion in parameter count, which grow linearly with the size of the number of tasks. Moreover, due to entangling knowledge across tasks, the second method suffers during few-shot adaptation to new tasks as it may overfit the training task distribution.

\begin{figure}
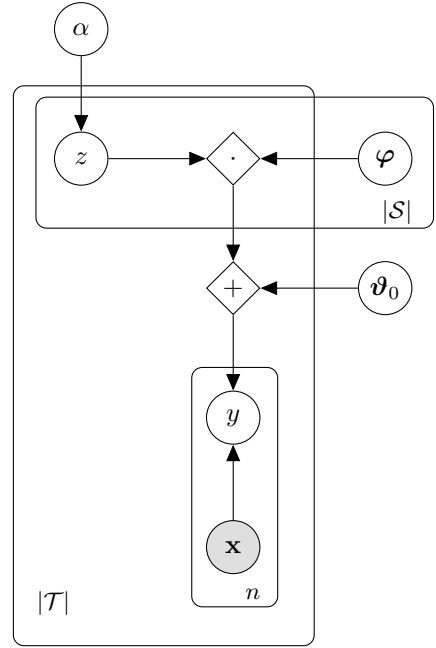

  \centering
  
  \tikz{ %
      \node[latent] (x) {$y$};%
      \node[obs,below=of x] (y) {$\xx$}; %
      \node[det,above=of x] (plus) {$+$}; %
      \node[latent,above=of plus,xshift=-2cm] (zeta) {$z$}; %
      \node[latent,above=of x,xshift=2cm] (base) {$\vtheta_0$}; %
      \node[det,above=of plus] (dot) {$\cdot$}; %
     \node[latent,above=of zeta] (alpha) {$\alpha$}; %
     \node[latent,above=of plus,xshift=2cm] (modules) {$\vphi$}; %
     \plate [inner sep=.2cm,xshift=.02cm,yshift=.1cm] {plate1} {(x) (y)} {$n$}; %
     \plate [inner sep=.25cm,xshift=.02cm,yshift=.2cm] {plate1} {(zeta) (modules)} {$|\mathcal{S}|$}; %
     \tikzset{plate caption/.append style={below=6pt and 0pt of #1.south west}}
     \plate [inner sep=.5cm,xshift=.2cm,yshift=.1cm] {plate1} {(x) (y) (zeta)} {$|\mathcal{T}|$}; %

     \edge {y} {x}
     \edge {alpha} {zeta}
     \edge {modules,zeta} {dot}
     \edge {dot} {plus}
     \edge {plus} {x}
     \edge {base} {plus}
  }

  \caption{A graph (plate notation) of the generative model of neural modules. Shaded circles refer to observed variables.}
  \label{fig:grf}
\end{figure}

In this work, we posit instead that there exists a (possibly small) fixed inventory of skills $\calS = (\calS_1, \dots, \calS_{|\calS|})$, where $|\calS| \ll |\calT|$. Each skill is an independent facet of knowledge that is reused across a subset of tasks. These assumptions guarantee both scalability and modularity. In particular, we seek to create a model that jointly learns which skills are active for which task, aggregates the corresponding skill parameters according to some deterministic function, and maximises the multitask log-likelihood
\begin{equation}
\label{eq:model}
\sum_{\mathcal{T}_i}\sum_{(\xx, y) \in \mathcal{T}_i} \log p(y \mid \xx, Z, \Phi, \calT_i) \, p(Z \mid \alpha) \, p(\Phi)
\end{equation}
with respect to two distinct sets of parameters: i) a matrix $Z$ of soft partitions of skills across tasks, regularised by a prior with hyper-parameter $\alpha$; ii) a matrix $\Phi$ of skill-specific parameters for a given neural architecture, which are composed according to the active skills. The full graphical model is shown in \cref{fig:grf}. In what follows, we illustrate each component separately.

\subsection{Soft Partitions}
What is the best strategy to determine which skills are active for which task?
The cognitively inspired notion of modularity at the level of structured inputs assumes that modules compete with each other for activation and updating~\citep{bengio2017consciousness,goyal2019recurrent}. This intuition is translated in practice into a softmax across modules and top-\textit{k} selection. Instead, we argue, modularity at the task level should reflect the fact that tasks fall into a hierarchy where more complex ones subsume simpler ones (e.g., dialogue requires both intention classification and text generation). Hence, variable-size subsets of skills should be allowed.

As a consequence, we assume that the matrix $Z \in \{0, 1\}^{|\calT| \times |\calS|}$ representing task--skill allocations is a soft partition: each cell $z_{ij}$ is a binary scalar indicating if module $\vphi_j$ is active for a certain task $\calT_i$.
However, being discrete, such a binary matrix is not differentiable and therefore cannot be learned end-to-end via gradient descent. Instead, we implement it as a collection of continuously relaxed Bernoulli distributions through a Gumbel-sigmoid \citep{maddison2016concrete,jang2016categorical}, which ensures stochasticity while allowing for differentiable sampling:
\begin{align} \label{eq:allocation}
    \hat{z}_{i,j} &=  \sigma \left[ \log \frac{\sigma({z}_{i,j}) \, u}{(1 - \sigma({z}_{i,j})) \, (1 - u)}^{1/\tau}\right] \nonumber \\ u &\sim \mathrm{Uniform}(0, 1).
\end{align}
In principle, either a coarse-grained soft partition can be learned globally for the entire neural network, or different fine-grained soft partitions can be assigned to each layer. We opt for the second alternative as it affords the model more flexibility and, foreshadowing, yields superior performance. Therefore, $Z$ and $\Phi$ are henceforth assumed to be layer-specific, although we will omit layer indexes in the notation for simplicity's sake. 

\subsection{Skill-specific Parameters}
Given the matrix row $\hat{Z}_{i,\star}$ for task $\calT_i$ from \cref{eq:allocation} and a matrix of skill-specific parameters $\Phi \in \mathbb{R}^{|\calS| \times d}$, where $d$ is the dimension of the layer parameters, the aggregate of the parameters of active skills is superimposed to a base parameterisation $\vtheta_0 \in \mathbb{R}^d$ shared across tasks. For instance, $\vtheta_0$ may be either the initialisation from a pre-trained model or learned from scratch:
\begin{align} \label{eq:skillparams}
    p(y | \xx, Z, \Phi, \calT_i) \triangleq p(y | x, \vtheta_i) \nonumber \\
    \vtheta_i = \vtheta_0 + \sum_{\calS_j} \Phi_{j,\star} \frac{\hat{z}_{i, j}}{\sum_{\calS_j} \hat{z}_{i,j}}.
\end{align}
Note that we normalise the rows of the task--skill allocation matrix $\hat{Z}$ prior to composition because the variable number of active skills per task would otherwise affect the norm of the combined parameters $\vtheta_i$, thus making training unstable.

\subsection{Inductive Biases}
A possible failure mode during training is a collapse into a highly entropic or non-sparse allocation matrix $\hat{Z}$, where all skills are active and skill-specific parameters remain general-purpose rather than specialising. Thus, we also provide an inductive bias to encourage the model to learn a low-entropy, highly-sparse allocation matrix.

\paragraph{IBP Prior} A possible inductive bias consists of adding an Indian Buffet Process~\citep[IBP;][]{teh2007stick} prior to $Z$, as IBP is the natural prior for binary matrices representing soft partitions. In particular, this assumes that the $i$-th task is associated with skills used by previous tasks with probability $\mathrm{Bernoulli}(m_{\calS_j} / i)$ (i.e., in proportion to their `popularity') and $\mathrm{Poisson}(\alpha / i)$ new skills, where $m_{\calS_j} = \sum_{\calT_i} z_{i, j}$ is the count of tasks for which skill $\calS_j$ is active. The log-probability of a task--skill allocation matrix under an IBP prior with hyper-parameter $\alpha$ is:
\begin{equation} \label{eq:ibp}
\begin{aligned}
        \log p(Z \mid \alpha) &= |\calS| \log \alpha - \sum_{h=1}^{2^{|\calT|}-1} \log Z_h! \\ &- \alpha \sum_{\calT_i} H_{\calT_i}
        + \sum_{\calS_j} \bigl[ \log (|\calT| - m_{\calS_j})! \\  &+ \log (m_{\calS_j} - 1)! - \log (|\calT|)!  \bigr] 
\end{aligned}
\end{equation}
where $H_n$ is the $n$-th harmonic number and $Z_h$ is the number of skills possessing the history $h$ (the binary vector of their corresponding column in the matrix $Z$). While training a neural network, the prior probability in \cref{eq:ibp} can be taken into account in the form of a regulariser subtracted to the main loss function.

\paragraph{Two-speed Learning Rate} As a simpler inductive bias alternative to the IBP prior, we also experiment with setting the learning rate for $Z$ higher than for $\Phi$. Intuitively, by accelerating learning of the soft partition matrix, to minimise the loss it becomes more convenient to discover better task--skill allocations over settling for general-purpose parameters that are agnostic with respect to the subset of active skills.

\subsection{Parameter Efficiency} \label{ssec:parameff}
In order to keep the skills modular, each of them must correspond to a separate layer parameterisation. Nevertheless, this may lead to a significant increase in both time and space complexity. Thus, we explore parameter-efficient implementations of $\Phi$ that only add a negligible amount of parameters to the base model. In particular, we contemplate both sparse and low-rank approximations.

\paragraph{Sparse Approximations} Lottery Ticket Sparse Fine-Tuning \citep[LT-SFT;][]{ansell2021composable} learns a highly sparse vector of differences with respect to a base model $\vtheta_0$. In our setting, this amounts to identifying a binary matrix $M \in \{0, 1\}^{|\calS| \times d}$ indexing non-zero entries in $\Phi$.\footnote{We leverage PyTorch \texttt{sparse\_coo\_tensor} for implementation.} We infer $M$  by selecting the top-$k$ entries in $\Phi$ based on their change in magnitude after a few early episodes:
\begin{equation}
    \mu_{i, j} \gets \begin{cases} 1 & \mathrm{if} \; \phi_{i, j} \in \mathop{\mathrm{argmax}}_{\phi_1, \dots, \phi_k} |\Phi^\prime - \Phi| \\ 0 & \mathrm{otherwise} \end{cases}
\end{equation}
The advantage of LT-SFT over other methods is that it is architecture-agnostic. However, it suffers from high space complexity during the early phase of training.

\paragraph{Low-rank Approximations}
Other parameter-efficient methods are designed for Transformer architectures specifically. For instance, Low-Rank Adapters \citep{hu2021lora} factorise each weight of the linear projections inside self-attention layers as a multiplication between two low-rank matrices. Hence, a linear projection $f_\vphi\ : \mathbb{R}^i \rightarrow \mathbb{R}^o$ is implemented as:
\begin{equation} \label{eq:lora}
    \xx^\prime = [W_{0} + (\zz^\top A B)] \xx + \bb_0
\end{equation}
where $A \in \mathbb{R}^{|\calS| \times o \times r}$, $B \in \mathbb{R}^{|\calS| \times r \times i}$, and $r$ is the rank. Hence, $\Phi \triangleq \textrm{flatten}(AB)$.

\subsection{Baselines}
We measure the performance of our \skilled approach, where we learn the skill--task allocation matrix $Z$ end-to-end, against the following baselines:
\begin{itemize}[leftmargin=12pt]
\itemsep0pt
\item \private: there is a separate model parameterisation for each task. During few-shot adaptation, given that $\calT_{train} \cap \calT_{eval} = \emptyset$, this model cannot benefit from any transfer of information between training and evaluation tasks. This is equivalent to the special case where the task--skill allocation matrix is an identity matrix $Z = I$ of size ${|\calT| \times |\calT|}$ and $|\calS| = |\calT|$.
\item \shared: a shared skill is learnt on the training tasks and then fine-tuned for each evaluation task separately.
This is equivalent to the special case where the task--skill allocation matrix is a matrix of ones $Z = \mathbf{1}$ of size $|\calT| \times 1$ and $|\calS| = 1$.
\item \expert, where the task--skill allocation is contingent on expert knowledge about task relationships. Crucially, $Z$ is fixed \textit{a priori} rather than being learned.
\end{itemize}
\noindent
In addition, we compare our \skilled method to a state-of-the-art baseline for multitask learning where parameters are softly shared, \hyperformer \citep{mahabadi2021parameter}. This method takes inspiration from \citet{ponti2020parameter} and generates adapters for the pre-trained model parameters with hyper-networks conditioned on task embeddings. In particular, parameters for task $\calT_i$ are obtained as:
\begin{equation}
    \xx^\prime = [W_0 + f_A(\calT_i) \, f_B(\calT_i)] \xx + \bb_0
\end{equation}
where each $f_W(\calT_i) \triangleq W_2[\mathrm{ReLU}(W_1\mathrm{e}(\calT_i))]$ is a hyper-network, $\mathrm{e}(\calT_i)$ is the embedding of task $\calT_i$, $W_1 \in \mathbb{R}^{h \times e}$ and $W_2 \in \mathbb{R}^{d \times h}$ for hidden size $h$ and task embedding size $e$. Note that, in our case, we generate LoRA rather than Adapter layers, contrary the original formulation of \citet{mahabadi2021parameter}, in order to make the baseline comparable to the implementation of \skilled in \cref{eq:lora}.

\section{Reinforcement Learning Experiments}
\subsection{Dataset}

As a proof-of-concept experiment, we perform multitask reinforcement learning on the BabyAI platform \citep{chevalier-boisvert2018babyai}. This benchmark consists in a series of increasingly complex levels, where an agent must execute a linguistic command by navigating a two-dimensional grid world and manipulating objects. Crucially, levels are procedurally generated to reflect different subsets of skills (e.g., \textsc{PickUp}, \textsc{Unlock}, \dots). This enables us to test our model in a controlled setting where performance based on learned skills can be compared with `ground truth' skills. In particular, we focus on a similar subset of 8 levels as \citet{hui2020babyai}: \textsc{GoToObj}, \textsc{GoToRedBallGrey}, \textsc{GoToRedBall}, \textsc{GoToLocal}, \textsc{PutNextLocal}, \textsc{PickupLoc}, \textsc{GoToObjMaze}, \textsc{GoTo}.

During each episode within a level, the visual input is a 7 by 7 grid of tiles, a partial observation of the environment based on the agent's field of view at the current time step. Each tile consists of 3 integers corresponding to the properties of a grid cell: object type, colour, and (optionally for doors) if they are open, closed, or locked. The linguistic instructions in English may require to complete multiple goals (via coordination) or express a sequence of sub-goals (via temporal markers).

\subsection{Experimental Setup}

\paragraph{Model Architecture}
The neural architecture adheres to the best model reported in \citet{hui2020babyai}, \textsc{bow\_endpool\_res}. It encodes the linguistic input through a Gated Recurrent Unit \citep[GRU;][]{cho2014properties} and the visual input through a convolutional network (CNN). These two streams from different modalities are then merged into a single representation through FiLM \citep{perez2018film}. This component performs a feature-wise affine transformation of the CNN output conditioned on the GRU output.

Afterwards, a Long Short-Term Memory network \citep[LSTM;][]{hochreiter1997long}, a recurrent module keeping track of the agent state trajectory, receives the multimodal representation and returns the current hidden state. This in turn is fed into two distinct MLPs, an actor and a critic \citep{sutton1984temporal}. The actor yields a distribution over actions, whereas the critic a reward baseline for the current state. In our experiments, each row of the matrix $\Phi$ corresponds to a possible parameterisation for all these components.

To determine \textit{a priori} a skill--task allocation for the \expert baseline, we harness the information about the skills employed to procedurally generate each level by \citet[p.\ 6]{chevalier-boisvert2018babyai}, which are indicated in \cref{tab:babyai_task_allocation}. For the \skilled model, we set $|\calS| = 9$ similarly to \expert. This allows us to compare learned skills and `ground-truth' skills from an inventory of identical size. As a parameter-efficient implementation of $\Phi$, we employ LT-SFT \citep{ansell2021composable} with a sparsity of $90\%$. For all model variants, $\vtheta_0$ and $\Phi$ are both initialised from a Kaiming uniform and learnable. During training, we sample levels uniformly.

\paragraph{Hyper-parameters}
We follow closely the best hyper-parameter setup of \citet{hui2020babyai}. Tiles in the visual input are encoded into embeddings of size 128 via a look-up table.
The CNN has 2 layers, filter size 3, stride 1, and padding 1; whereas the FiLM module has 2 layers. A residual layer is added between the CNN output and each of the FiLM layers. The output of FiLM is max-pooled with a layer of size 7 and stride 2. Both the LSTM and the GRU have a hidden size of 128.

A learning rate of 1e-4 is adopted for Adam \citep{kingma2014adam}. We optimise the model with Proximal Policy Optimisation \citep[PPO;][]{schulman2017proximal} and Back-Propagation Through Time \citep[BPTT;][]{werbos1990backpropagation}. Additionally, we use an Advantage Actor--Critic \citep[A2C;][]{wu2017scalable} with Generalised Advantage Estimation \citep[GAE;][]{schulman2015high}. The reward is calculated as $(1 - 0.9 \, n / n_{\max})$ if the agent completes a task---where $n$ is the number of steps required and $n_{\max}$ is a threshold set according to the level difficulty, $0$ otherwise. Returns are discounted by $\gamma$ = 0.99.

\subsection{Results}

We now measure whether our latent-skill model facilitates sample efficiency, which following \citet{chevalier-boisvert2018babyai} is defined as the number of episodes required for an agent to reach a success rate greater than $0.99$. Success in turn is defined as executing an instruction in a number of steps $n < n_{\max}$, where the threshold again depends on the level complexity.

\begin{figure}[t]
    \centering
    \includegraphics[width=\columnwidth]{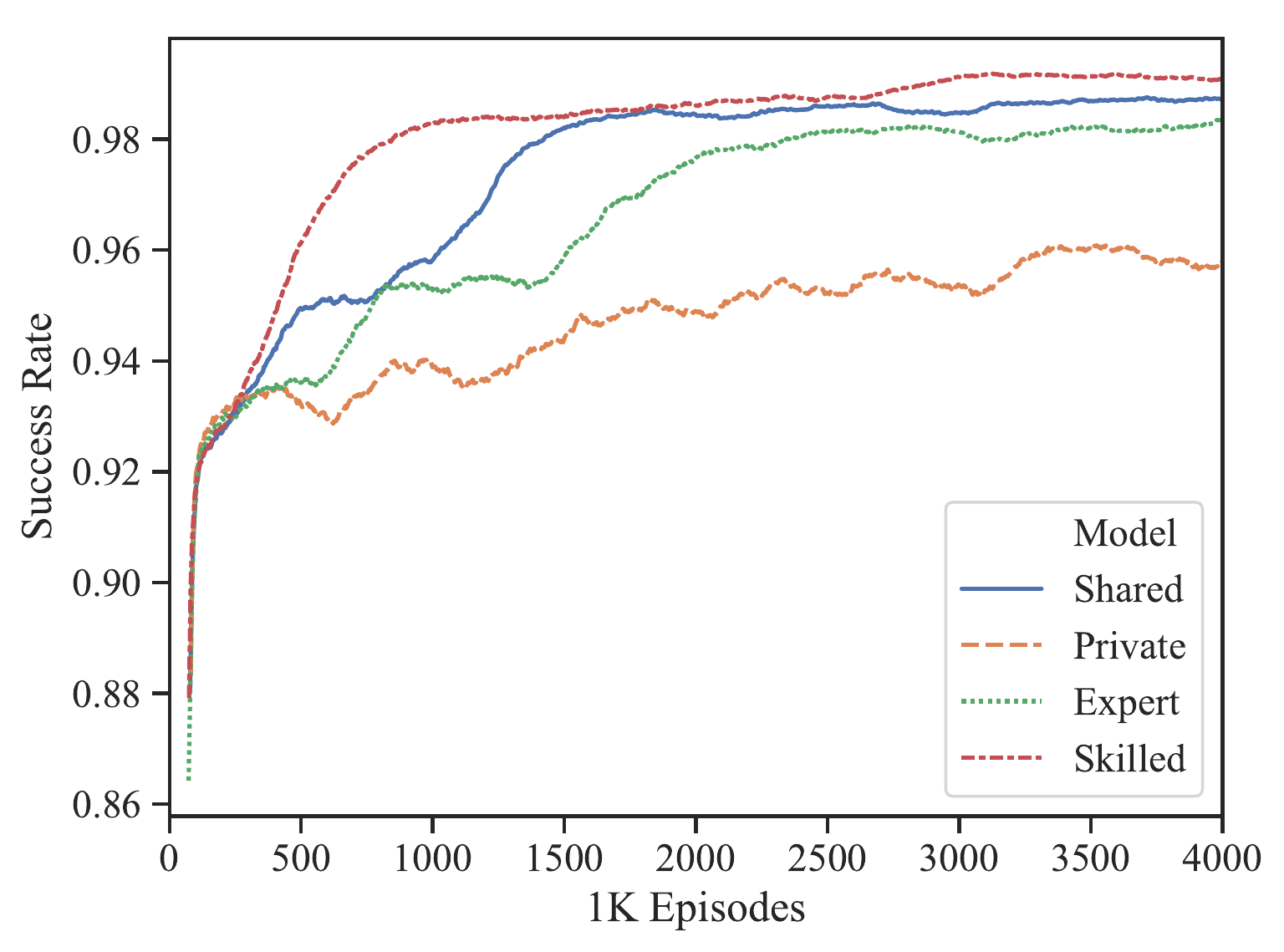}
    \caption{Smoothed sample efficiency (success rate vs.\ number of episodes) for different multitask models across 8 levels of BabyAI.}
    \label{fig:sampleffbabyai}
\end{figure}

We plot our results in \cref{fig:sampleffbabyai}. Firstly, models sharing information across tasks (either fully or mediated by skills) enjoy higher sample efficiency than assigning disjoint parameters for each task (\private), as they can borrow statistical strength from each other. Crucially, among information-sharing models, \skilled (where knowledge is modular) surpasses \shared (where knowledge is entangled among tasks). Thus, considering a task as a collection of fine-grained skills that can be separated and reused is the most effective way of sharing information. Finally, results surprisingly show that learning a task--skill allocation matrix end-to-end (\skilled) is more beneficial than leveraging the ground-truth task--skill decomposition used to create the BabyAI levels (\expert). This highlights the fact that different tasks might mutually benefit in ways that go beyond what is posited \textit{a priori} by experts, and that our proposed approach can successfully uncover and exploit such task synergies.

Moreover, we run several ablations to study the impact of the inductive biases and the parameter-efficient implementation. We report the number of episodes required to reach a success rate $> 0.99$ in \cref{tab:rlsampleff}, comparing the \skilled model in the standard setup (with two-speed learning rates and sparse skill-specific parameters) with other variants (with an IBP prior or fully dense skill-specific parameters). We find that adding an IBP prior with $\alpha=5$ does not affect the performance significantly. Hence, for the remainder of the experiments, we will adopt two-speed learning rates as an inductive bias. On the other hand, employing fully dense skill-specific parameters increases sample efficiency, albeit to a limited degree. Thus, we verify that parameter sparsification is an effective trade-off between performance and space complexity.

\begin{table*}[t]
    \footnotesize
\begin{tabularx}{\textwidth}{>{\sc \scriptsize}l >{\scriptsize}l|P|PPPPP}

\toprule
                                           Task &               Metric & \citet{ye-etal-2021-crossfit} &                        Shared &                       Private &                        Expert &                   HyperFormer &                      Skilled \\
\midrule
                                        ag-news &    C--F1 &   84.6 $\pm$  \hspace{4pt}1.4 &               59.6 $\pm$ 21.1 &               74.7 $\pm$ 10.2 &   47.0 $\pm$  \hspace{4pt}9.9 &               64.6 $\pm$ 13.4 &  81.2 $\pm$  \hspace{4pt}8.0 \\
                                        ai2-arc &                  Acc &   22.8 $\pm$  \hspace{4pt}1.9 &   23.7 $\pm$  \hspace{4pt}2.4 &   20.3 $\pm$  \hspace{4pt}0.8 &   28.3 $\pm$  \hspace{4pt}3.7 &   23.8 $\pm$  \hspace{4pt}6.7 &  22.3 $\pm$  \hspace{4pt}3.4 \\
                                amazon-polarity &    C--F1 &   92.2 $\pm$  \hspace{4pt}0.6 &   92.4 $\pm$  \hspace{4pt}2.8 &   94.4 $\pm$  \hspace{4pt}0.4 &   57.4 $\pm$  \hspace{4pt}3.1 &   93.8 $\pm$  \hspace{4pt}1.5 &  93.7 $\pm$  \hspace{4pt}0.9 \\
 bsn-npi-licensor &                  Acc &   99.9 $\pm$  \hspace{4pt}0.2 &  99.9 $\pm$  \hspace{4pt}0.0 &  99.9 $\pm$  \hspace{4pt}0.0 &  99.9 $\pm$  \hspace{4pt}0.0 &  99.9 $\pm$  \hspace{4pt}0.0 &  99.9 $\pm$  \hspace{4pt}0.2 \\
            bsn-npi-scope &                  Acc &               85.7 $\pm$ 13.0 &   99.8 $\pm$  \hspace{4pt}0.3 &   99.6 $\pm$  \hspace{4pt}0.9 &   99.9 $\pm$  \hspace{4pt}0.2 &  99.9 $\pm$  \hspace{4pt}0.0 &  99.9 $\pm$  \hspace{4pt}0.2 \\
                                     break-QDMR &                   EM &    4.8 $\pm$  \hspace{4pt}0.4 &    4.1 $\pm$  \hspace{4pt}0.5 &    1.9 $\pm$  \hspace{4pt}1.7 &    4.1 $\pm$  \hspace{4pt}0.9 &    3.8 $\pm$  \hspace{4pt}1.2 &   4.9 $\pm$  \hspace{4pt}0.6 \\
                                          circa &    C--F1 &   42.3 $\pm$  \hspace{4pt}7.8 &   44.3 $\pm$  \hspace{4pt}7.5 &   22.2 $\pm$  \hspace{4pt}6.3 &   13.0 $\pm$  \hspace{4pt}7.0 &   25.0 $\pm$  \hspace{4pt}6.3 &  45.9 $\pm$  \hspace{4pt}5.7 \\
                                   crawl-domain &                   EM &   20.7 $\pm$  \hspace{4pt}2.0 &   40.9 $\pm$  \hspace{4pt}2.2 &   36.2 $\pm$  \hspace{4pt}6.3 &   37.1 $\pm$  \hspace{4pt}5.3 &   34.9 $\pm$  \hspace{4pt}3.8 &  39.0 $\pm$  \hspace{4pt}4.2 \\
                               ethos-disability &    C--F1 &   75.3 $\pm$  \hspace{4pt}2.2 &               66.9 $\pm$ 11.8 &               64.7 $\pm$ 12.1 &               59.2 $\pm$ 15.8 &   79.4 $\pm$  \hspace{4pt}3.9 &  72.2 $\pm$  \hspace{4pt}5.2 \\
                       ethos-sexual &    C--F1 &   59.7 $\pm$  \hspace{4pt}5.7 &   62.4 $\pm$  \hspace{4pt}8.4 &   71.2 $\pm$  \hspace{4pt}9.8 &               40.3 $\pm$ 11.4 &  76.8 $\pm$ 10.0 &  86.1 $\pm$  \hspace{4pt}2.6 \\
                                    freebase-qa &                   EM &    1.3 $\pm$  \hspace{4pt}0.1 &    0.7 $\pm$  \hspace{4pt}0.2 &    0.2 $\pm$  \hspace{4pt}0.1 &    1.3 $\pm$  \hspace{4pt}0.5 &    1.6 $\pm$  \hspace{4pt}0.8 &   0.7 $\pm$  \hspace{4pt}0.3 \\
                                      glue-cola &  M-Corr &    3.5 $\pm$  \hspace{4pt}6.7 &   12.9 $\pm$  \hspace{4pt}5.5 &    9.1 $\pm$  \hspace{4pt}6.3 &    7.6 $\pm$  \hspace{4pt}5.6 &    6.8 $\pm$  \hspace{4pt}4.2 &   7.1 $\pm$  \hspace{4pt}5.3 \\
                                      glue-qnli &                  Acc &   74.7 $\pm$  \hspace{4pt}2.9 &   75.5 $\pm$  \hspace{4pt}3.6 &   57.1 $\pm$  \hspace{4pt}7.7 &               56.6 $\pm$ 19.7 &   73.9 $\pm$  \hspace{4pt}3.2 &  78.1 $\pm$  \hspace{4pt}1.6 \\
                                     hatexplain &    C--F1 &   44.9 $\pm$  \hspace{4pt}2.5 &   33.1 $\pm$  \hspace{4pt}8.2 &   26.5 $\pm$  \hspace{4pt}7.8 &   11.9 $\pm$  \hspace{4pt}4.0 &   23.2 $\pm$  \hspace{4pt}6.2 &              32.6 $\pm$ 13.6 \\
                                         quoref &                QA-F1 &   41.2 $\pm$  \hspace{4pt}1.6 &   46.0 $\pm$  \hspace{4pt}4.4 &   36.3 $\pm$  \hspace{4pt}4.6 &   48.4 $\pm$  \hspace{4pt}4.3 &   41.7 $\pm$  \hspace{4pt}6.5 &  47.3 $\pm$  \hspace{4pt}3.5 \\
                                      race-high &                  Acc &   30.5 $\pm$  \hspace{4pt}1.5 &   34.0 $\pm$  \hspace{4pt}2.7 &   28.5 $\pm$  \hspace{4pt}1.4 &   38.5 $\pm$  \hspace{4pt}2.0 &   30.8 $\pm$  \hspace{4pt}1.9 &  34.8 $\pm$  \hspace{4pt}2.0 \\
                                  superglue-rte &                  Acc &   60.4 $\pm$  \hspace{4pt}3.6 &   60.6 $\pm$  \hspace{4pt}2.9 &   49.7 $\pm$  \hspace{4pt}5.1 &   51.7 $\pm$  \hspace{4pt}4.8 &   60.9 $\pm$  \hspace{4pt}3.8 &  60.4 $\pm$  \hspace{4pt}5.9 \\
                               tweet-eval-irony &    C--F1 &   55.2 $\pm$  \hspace{4pt}3.6 &   52.1 $\pm$  \hspace{4pt}8.0 &               50.1 $\pm$ 14.2 &   25.6 $\pm$  \hspace{4pt}8.9 &   38.4 $\pm$  \hspace{4pt}6.0 &  57.2 $\pm$  \hspace{4pt}2.4 \\
                                     wiki-split &              Rouge-L &   79.3 $\pm$  \hspace{4pt}0.5 &   80.1 $\pm$  \hspace{4pt}0.6 &   80.3 $\pm$  \hspace{4pt}0.6 &   79.2 $\pm$  \hspace{4pt}0.8 &   79.2 $\pm$  \hspace{4pt}0.7 &  80.6 $\pm$  \hspace{4pt}0.3 \\
                                  yelp-polarity &    C--F1 &               71.8 $\pm$ 21.1 &               88.3 $\pm$ 14.9 &               65.0 $\pm$ 20.5 &               53.9 $\pm$ 12.7 &   95.0 $\pm$  \hspace{4pt}0.9 &  94.5 $\pm$  \hspace{4pt}1.1 \\
                                  \hline
                                            all &                  Avg &               52.5 $\pm$ 30.8 &               53.9 $\pm$ 30.5 &               49.4 $\pm$ 31.7 &               43.0 $\pm$ 29.0 &               52.7 $\pm$ 33.2 &              \textbf{56.9} $\pm$ \textbf{32.3} \\
\bottomrule

\end{tabularx}

    \caption{Few-shot adaptation results of \skilled and four baselines, as well as the original model from \citet{ye-etal-2021-crossfit}. Results are reported both in aggregate and separately for each task in $\calT_{eval}$. For the full name of the metrics, refer to \cref{ssec:sldata}.}
    \label{tab:ft}
    \vspace{-4mm}
\end{table*}

\section{Supervised Learning Experiments}

\subsection{Dataset} \label{ssec:sldata}
In order to measure the benefits of a modular design for systematic generalisation to new tasks, we run a second set of experiments on CrossFit \citep{ye-etal-2021-crossfit}, a benchmark including 160 diverse natural language processing tasks sourced from Huggingface Datasets \citep{lhoest2021datasets}. The tasks in CrossFit are all converted into a unified text-to-text format inspired by \citet{raffel2020exploring}. Moreover, they are partitioned into three disjoint subsets. First, a model is pre-trained in a multitask fashion on training tasks $\calT_{train}$. Afterwards, it is adapted to each evaluation task from $\calT_{eval}$ in a few-shot learning setting. Hyper-parameter are tuned on the held-out set $\calT_{dev}$.

We adopt the partition 1 (called \textsc{Random}) from \citet{ye-etal-2021-crossfit}, where $|\calT_{train}| = 120$, $|\calT_{dev}| = |\calT_{eval}| = 20$, and tasks are split randomly. This is the most comprehensive partition and most suited for general-purpose models, as it includes all types of tasks. Every task is associated with 5 different few-shot data splits for train $\calD_{train}$ and development $\calD_{dev}$ and 1 larger data split for evaluation $\calD_{eval}$. During multitask pre-training, we concatenate all train and development splits of datasets from $\calT_{train}$, whereas during few-shot adaptation we use the splits of datasets in $\calT_{eval}$ separately in 5 distinct runs. We measure the performance of a model with 7 evaluation metrics according to the type of task: C[lassification]-F1, Acc[uracy], QA-F1, E[xact] M[atch], Rouge-L, M[atthew]-Corr[elation], and P[earson]-Corr[elation].

\begin{figure*}[t]
    \centering
    \includegraphics[width=\textwidth]{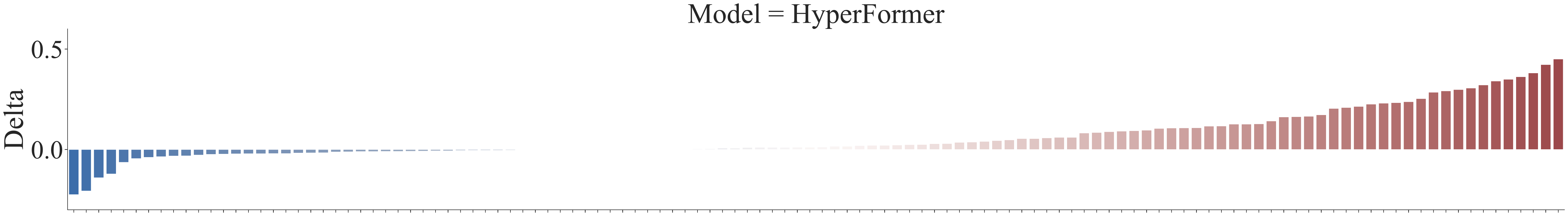}
    \par\vspace{2mm}
    \includegraphics[width=\textwidth]{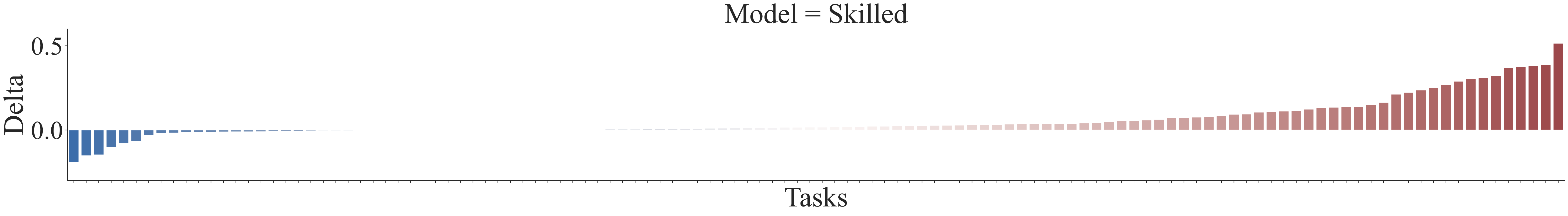}
    \caption{Delta in performance in terms of task-specific metrics between \skilled (top) and \hyperformer (bottom) on the one hand and \shared on the other, across 120 seen CrossFit tasks after multitask pre-training.}
    \label{fig:mlt_bar}
     \vspace{-2mm}
\end{figure*}

\subsection{Experimental Setup}

\paragraph{Model Architecture}
As pre-trained weights $\vtheta_0$ for conditional text generation, we choose BART Large \citep{lewis-etal-2020-bart}, a 24-layer Transformer-based encoder--decoder. We use LoRA \citep{hu2021lora} to implement skill-specific parameters efficiently, as it was explicitly designed for the Transformer architecture and has lower space complexity that LT-SFT during the early phases of training (cf.\ \cref{ssec:parameff}). While $\vtheta_0$ remains frozen for all model variants, $A$ matrices in $\Phi$ (and $f_A(\cdot)$ in \hyperformer) are initialised to zero matrices following \citet{hu2021lora}.

As a source of expert knowledge for the \expert baseline, we associate each task to a unique skill corresponding to one of the 4 task types of \citet[p.\ 20]{ye-etal-2021-crossfit}'s taxonomy: question answering, conditional text generation, classification, or other (e.g., regression). The skill inventory size for \skilled was chosen among $\{2, 4, 8, 16, 32\}$ based on validation. We set the embedding size \textit{e} and hidden size \textit{h} of \hyperformer to an identical value to ensure a fair comparison.

\paragraph{Hyper-parameters}

During both multitask pre-training and few-shot adaptation, we use the Adam optimiser \citep{kingma2014adam} and select a learning rate for $\Phi$ among $\{1e-2, 1e-3, 1e-4\}$ based on performance on the development set of $\calT_{train}$ and $\calT_{dev}$, respectively. As a more aggressive learning rate for $Z$ in \skilled instead we search in the range $[1e-1, 1e-2]$. We run multitask pre-training for 30 epochs with an effective batch size of 32, with a warm-up of 6\% of the total steps. Instead, during few-shot adaptation, the effective batch size is 8 for 1000 training steps, with a warm-up rate of 10\% and a weight decay of 1e-2.

For a comparison of parameter counts, LoRA adds $4l(2hr + |\calT|) \cdot |\calS|$ parameters to the pre-trained model, where $l$ is the number of layers in the encoder and decoder, $h$ is the hidden size, and $4$ is the number of linear projections in self-attention (query, key, value, output). We use $r=16$, $l=24$ and $h=1024$ for BART Large,  so we add $\sim 3 \cdot 10^6$ parameters per skill. Given that the pre-trained model has $\sim 4 \cdot 10^8$ parameters, this implies an increase of $\sim 0.78\%$ per skill. \hyperformer adds $4l(2he + 2e) \cdot e$ parameters, an increase of $\sim 0.78\%$ per task embedding dimension.

\subsection{Results}

\paragraph{Few-shot Adaptation to New Tasks}
In the few-shot adaptation setting, our goal is to evaluate the capability of the model to quickly generalise to new tasks unseen during training. This is the most realistic setting as tasks encountered by models deployed `in the wild' will be characterised by different distributions or involve different input / output spaces. Performance in terms of task-specific metrics is reported in \cref{tab:ft} for the 20 evaluation tasks individually and on average.  

Crucially, results show that \skilled outperforms alternative formulations of the task--skill allocation matrix, such as \shared, \private and \expert. Importantly, we note that \skilled also surpasses \hyperformer by a sizeable margin despite the two models having comparable parameter counts. This points to the fact that explicitly modularising knowledge learnt during multitask training is important for systematic adaptation to unseen tasks, whereas entangled knowledge is more brittle to distribution shifts in cross-task transfer. Finally, we corroborate the soundness of our experimental setup by reproducing the results of~\citet{ye-etal-2021-crossfit}.\footnote{Note that in \citet{ye-etal-2021-crossfit} the pre-trained model is BART Small and all parameters are fine-tuned. Hence, these results are not directly comparable.}

\paragraph{Multitask Evaluation on Seen Tasks}
Moreover, to ensure that modularity does not adversely affect in-domain performance, we evaluate models on the test sets of seen tasks after multitask training. The results are shown in \cref{tab:multitask}. \skilled achieves average improvements of up to \textasciitilde3\% in the global metrics with respect to \shared by flexibly allocating skills to each task. In Figure~\ref{fig:mlt_bar}, we report the delta in performance in terms of task-specific metrics between \skilled and \shared for the 120 training tasks, in most of which \skilled yields positive gains. In this context, we can see that \skilled achieves comparable performance to \hyperformer, therefore confirming that explicit modularisation can be as effective as conditional parameter generation when evaluated on seen tasks, but also engenders vast improvements on held-out tasks.

\paragraph{In-Depth Analysis of Learned Skills}
Finally, we run an in-depth analysis of the task--skill allocation matrices $Z$ learned by \skilled. Specifically, we measure: 
\begin{enumerate}[leftmargin=12pt]
\itemsep0pt
    \item \textit{Discreteness}. How close is the continuous relaxation to a binary matrix? To this end, we report the average normalised entropy across all probabilities in the cells of the matrix:
    \begin{equation}
    \mathrm{Discrete}(Z) = \frac{1}{|\calT| \cdot |\calS|} \sum_{\calT_i} \sum_{\calS_j} \frac{\mathcal{H}(z_{ij})}{\log 2}
    \end{equation}
    \item \textit{Sparsity}. How many skills are active per task on average? We count the rate of non-zero cells in the values rounded to the closest integer:
    \begin{equation}
    \mathrm{Sparsity}(Z) = \frac{1}{|\calT| \cdot |\calS|} \sum_{\calT_i} \sum_{\calS_j} \lfloor {z}_{ij} \rceil
    \end{equation}
    \item \textit{Usage}. Is the allocation of skills across tasks balanced or are some preferred over others? We provide the normalised entropy of a categorical distribution parameterised by $\sum_j Z_{\star, j}$, the sum of the columns of $Z$:
    \begin{equation}
    \mathrm{Usage}(Z) = \frac{\mathcal{H} \left[ \sum_{\calT_i} z_{i, \star} \right]}{ \log |\calS|}
    \end{equation}
\end{enumerate}
Note that the entropy values are normalised into the range $[0, 1]$ to make them invariant to the number of skills: this quantity is known as `efficiency'.

\begin{figure}[t]
    \centering
    \includegraphics[width=\columnwidth]{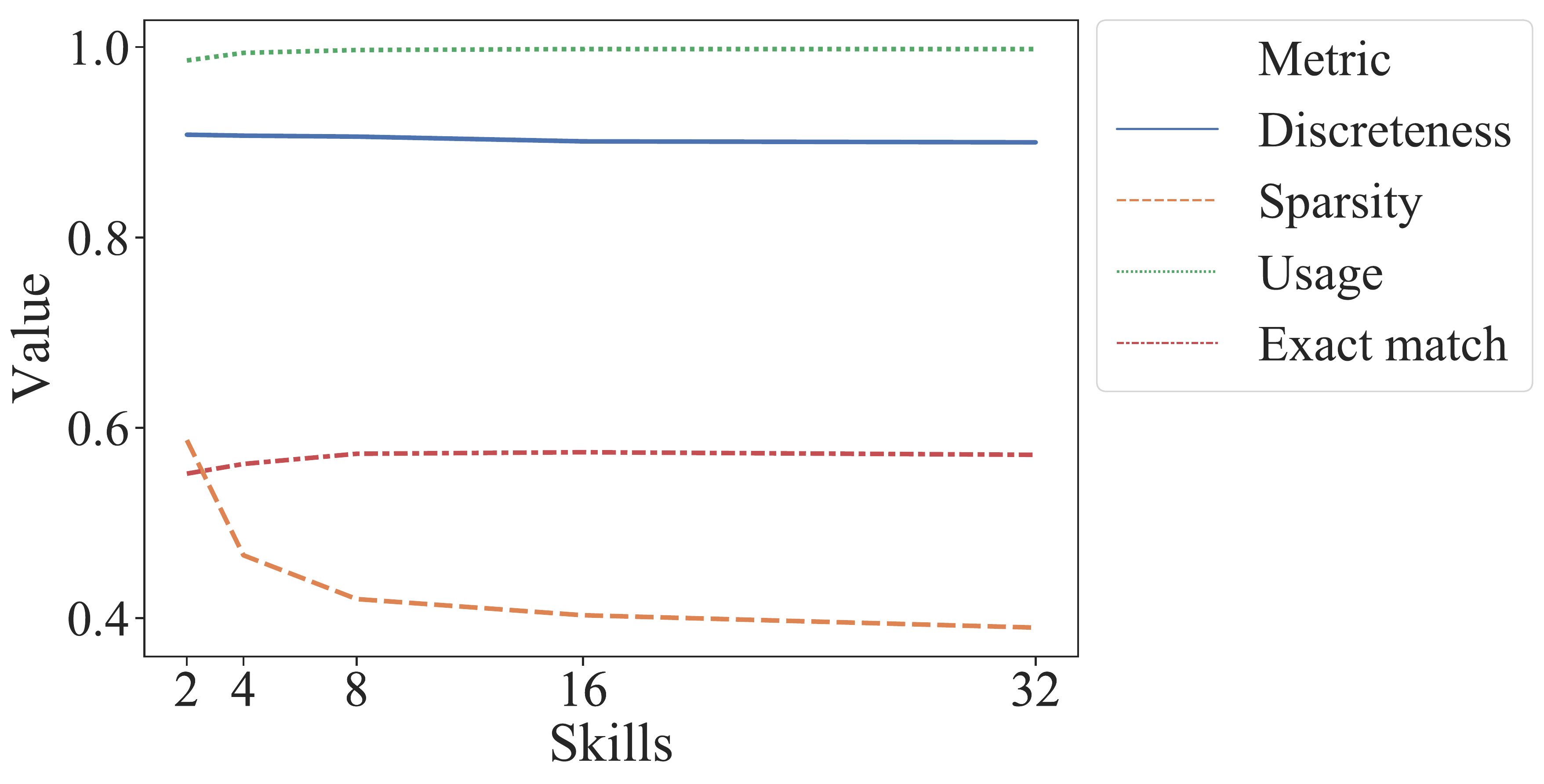}
    \caption{Statistics of the task--skill matrices for different choices of skill inventory size, including: discreteness, sparsity, usage, and the average exact match on the development set of 120 CrossFit tasks.}
    \label{fig:skillsearch}
\end{figure}

We plot these metrics---as well as the performance on in-domain train tasks in terms of exact match---as a function of the skill inventory size in \cref{fig:skillsearch}. We find that, whilst a continuous relaxation, the learned matrices are highly discretised and all their values are extremely close to either 0 or 1. Moreover, the level of sparsity decreases as the number of skills increases. This means that smaller subsets of skills are required in proportion due to the diversity of available skills. Finally, usage is consistently near the maximum value, which implies that there is uniformity in how frequently each skill is active across tasks.

\begin{figure}[t]
    \centering
    \includegraphics[width=\columnwidth]{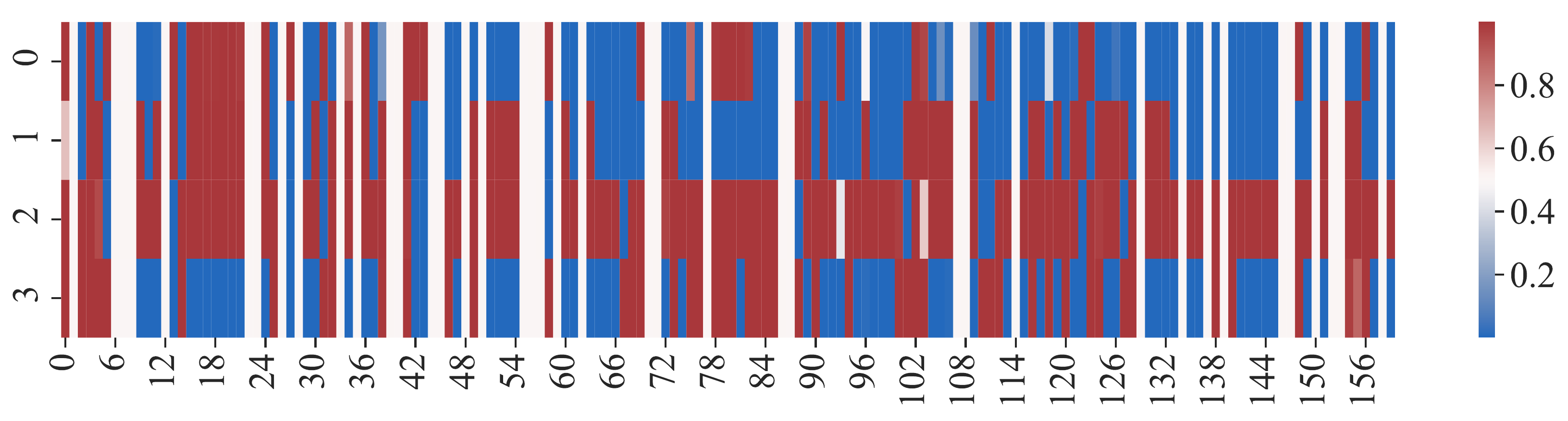}
    \caption{Posterior over $Z$ in \skilled for $|\mathcal{S}| = 4$.}
    \label{fig:Z_s4}
\end{figure}

\begin{figure}[t]
    \centering
    \includegraphics[width=\columnwidth]{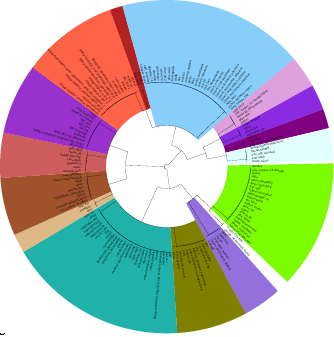}
    \caption{Task partitions for $|\mathcal{S}| = 4$, which corresponds to $2^{|\mathcal{S}|} = 16$ possible subsets of skills.}
    \label{fig:dendro}
\end{figure}

Overall, these results demonstrate that a quasi-binary, highly-sparse, and non-trivial allocation matrix can be successfully learned in an end-to-end fashion even with simple inductive biases such as a two-speed learning rate. For instance, we visualise the posterior of $Z$ for $|\calS| = 4$ in \cref{fig:Z_s4}. Crucially, the learned allocation also facilitates the interpretability of black-box multitask models. In fact, the structure of $Z$ corresponds to an explicit hierarchy of tasks, where simpler ones are subsumed by more complex ones, and similar tasks can be grouped into the same category if they share the same subset of skills. We plot this hierarchy as a dendrogram in \cref{fig:dendro}. For instance, most \textsc{GLUE} tasks \citep{wang-etal-2018-glue} are grouped together as they are all focused on natural language understanding: for instance, they require skill 1 (\textsc{cola}), 2 (\textsc{mrpc, rte, sst2, wnli}), or both 1 and 2 (\textsc{mnli, qqp}).

\section{Related Work}

\paragraph{Modular Networks}
The idea of modularising neural network computation by decomposing it into a subset of specialised sub-systems has long been sought as a way to achieve better generalisation to unseen inputs~\citep{jacobs1991adaptive,andreas2016nmn,kirsch2018modular}, tasks~\citep{jacobs1991task,alet2018modular,ruder2019latent} and recently to improve continual learning~\citep{ostapenko2021continual} and robustness to changes in the environment~\cite{goyal2019recurrent}. Modular networks fall in the category of conditional computation methods, where modules are chosen dynamically given the input~\citep{gulcehre2016dynamic}.
In routing networks~\citep{rosenbaum2019routing}, the system makes hard decisions about which modules to use and learns the structure in which modules are composed~\citep{andreas2016nmn, alet2018modular}.~\citet{andreas2016nmn} learn the structure using an external parser while~\citet{alet2018modular} recur to a stochastic process in which structures are sampled with simulated annealing. In mixture of experts (MoE) approaches, the system selects a potentially sparse, soft subset of modules depending on the input to be processed~\citep{jacobs1991adaptive,shazeer2017outrageously}. MoEs can be interpreted from the point of view of independent mechanisms~\citep{parascandolo2018learning} that~\citet{goyal2019recurrent} further extend to handle sequential problems. In the context of NLP,~\citet{fedus2021switch} successfully used a MoE architecture to scale large language model pre-training to trillions of parameters.

In contrast to previous approaches, our model conditions the computation on the task rather than on task inputs. There have been related attempts to enforce parameter reuse and modularity for multitask learning~\citep{rajendran2017adaapt,ponti2020parameter,kingetsu2021neural,kudugunta2021beyond}.~\citet{rajendran2017adaapt} learn separate modules for each task and then learn how to reuse those modules for a new task.~\citet{kudugunta2021beyond} uses a set of modules for each task in a multi-lingual translation setting. Our approach does not assume a set of modules for each task but instead decomposes a task into a set of skills themselves reusable across tasks.

\paragraph{Multitask NLP}
Multitask learning for NLP has been an effective strategy for improving model performance in low-resource tasks and for quickly adapting to new, unseen tasks~\cite{ruder2019latent,liu2019multi,min2021metaicl,wei2021finetuned,aribandi2021ext5,sanh2022multitask,mahabadi2021parameter,rusu2019meta}, languages \citep{ponti2019modeling}, and modalities \citep{bugliarello2022iglue}.~\citet{liu2019multi} adopt a multitask training strategy with a shared model and achieve impressive performance on GLUE. However, the method still requires task-specific fine-tuning. Rather than re-training all the model parameters,~\citet{houlsby2019parameter} proposes to train task-specific adapters.~\citet{pfeiffer2020adapterfusion} share information across task-specific adapters while alleviating negative task interference. Instead of using adapters, in our experiments we parameterise our skills with LT-SFT \citep{ansell2021composable} or LoRA~\citep{hu2021lora}, which achieve comparable or superior performance. Recently,~\citet{mahabadi2021parameter} ensure cross-task information sharing by using a hyper-network to generate task-specific adapters. Differently, our task-specific parameters are composed of a set of skills from a shared inventory, which makes our approach modular and more scalable.

\paragraph{Few-shot Task Adaptation} Several multitask approaches specifically target adaptation to new tasks, such as meta-learning approaches~\citep{alet2018modular, rusu2019meta,ponti-etal-2021-minimax, garcia2021meta,ostapenko2021continual}. In our paper, we efficiently achieve few-shot task adaptation by inferring the task--skill allocation matrix for new tasks and fine-tuning skill parameters, which were previously learned via multitask learning. In fact, \citet{ye-etal-2021-crossfit} found that this pre-training routine is superior to meta-learning in CrossFit. 
A similar attempt to recompose modular knowledge learnt on previous tasks has been recently explored by~\citet{ostapenko2021continual}.

\section{Conclusions}
In this work, we argued that a modular design is crucial to ensure that neural networks can learn from a few examples and generalise robustly across tasks by recombining autonomous facets of knowledge. To this end, we proposed a model where a subset of latent, discrete skills from a fixed inventory is allocated to each task in an end-to-end fashion. The task-specific instantiation of a neural network is then obtained by combining efficient parameterisations of the active skills, such as sparse or low-rank adapters. We evaluate the sample efficiency of our model on multitask instruction following through reinforcement learning and its few-shot adaptability on multitask text-to-text generation through supervised learning. In both experiments, we surpass competitive baselines where parameters are fully shared, task-specific, combined according to expert knowledge, or generated conditionally on the task. Finally, we show that our model facilitates interpretability by learning an explicit hierarchy of tasks based on the skills they require.

\bibliography{tacl2018}
\bibliographystyle{acl_natbib}

\clearpage
\appendix

\begin{figure*}[t]
    \centering

    \includegraphics[width=\textwidth]{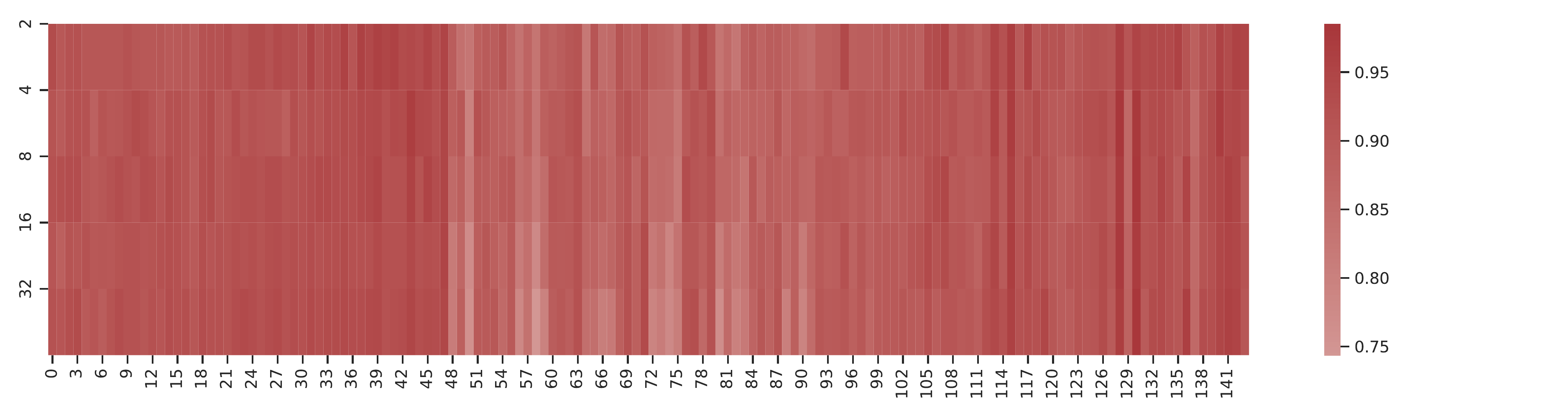}
    \includegraphics[width=\textwidth]{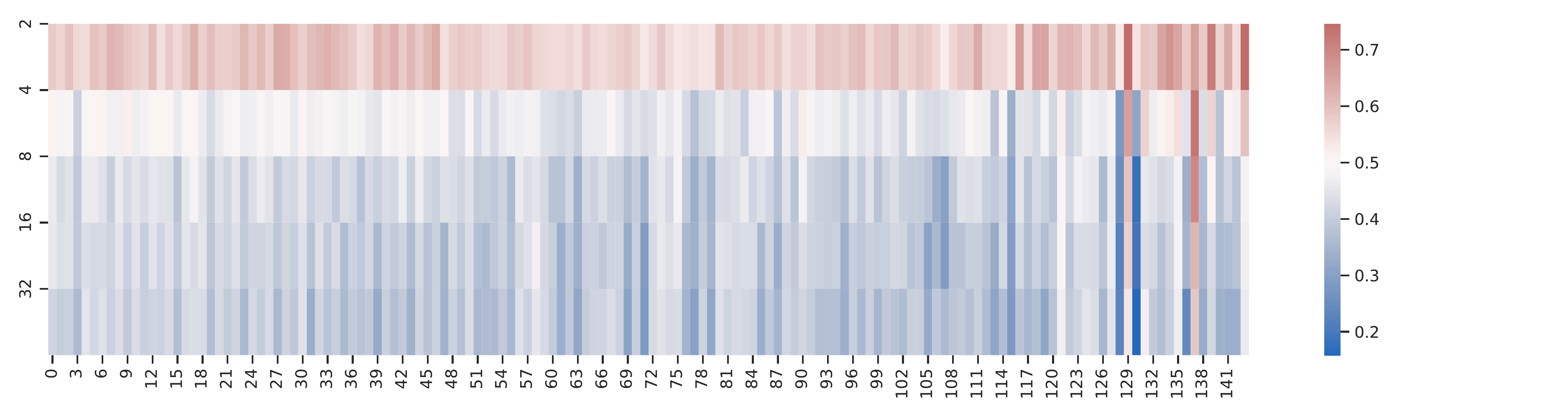}
    \caption{Per-layer discreteness (top) and per-layer sparsity (bottom).}
    \label{fig:discretesparse}
\end{figure*}

\section{Additional Results for BabyAI}

\begin{table}[h]
    \centering
    \begin{tabularx}{.75\columnwidth}{l|>{\sc}X}
\toprule
Skills & {\normalfont Level} \\
\midrule
        1 2 3 4 8 & GoTo \\
1 8 & GoToObjMaze \\
1 2 3 6 7 & PickupLoc \\
1 2 3 5 & PutNextLocal \\
1 2 3 4 & GoToLocal \\
1 2 3 & GoToRedBall \\
1 2 & GoToRedBallGrey \\
1 & GoToObj \\
\bottomrule
    \end{tabularx}
    \caption{BabiAI \expert task--skill allocation.}
    \label{tab:babyai_task_allocation}
\end{table}

\begin{table}[h]
    \centering
    \begin{tabular}{>{\sc}r|r}
    \toprule
    {\normalfont Model }& Episodes \\
    \midrule
    Private & >6000000 \\
    Shared & 3544294 \\
    Expert & 4608019 \\
    Skilled & \textbf{2218226}  \\
        \hline
        + IBP prior & 2143491 \\
        - sparsity & {1853060} \\
    \bottomrule
    \end{tabular}
    \caption{Sample efficiency of various models on 8 BabyAI levels measured as the number of episodes needed to reach a success rate $>0.99$.}
    \label{tab:rlsampleff}
\end{table}

\section{Additional Results for CrossFit}

\begin{table}[h]
    \centering
    \begin{tabularx}{.95\columnwidth}{rYYY}
    \toprule
Metric & \shared & \hyperformer & \skilled \\
    \midrule
    \rowcolor{Gray} \multicolumn{4}{c}{\textsc{Task-specific}}\\
Acc & 58.47 & 62.63 & 62.66 \\
C-F1 & 41.76 & 56.74 & 55.39 \\
EM & 20.05 & 21.68 & 21.70 \\
P-Corr & 56.83 & 61.54 & 52.60 \\
QA-F1 & 48,88 & 51.04 & 52.35 \\
Rouge-L & 28.02 & 26.71 & 28.05 \\
    \midrule
        \rowcolor{Gray} \multicolumn{4}{c}{\textsc{Global}}\\
Average & 43.09 & 49.37 & 48.95 \\
    \bottomrule
    \end{tabularx}
    \caption{Performance of multitask models averaged over test sets of 120 seen CrossFit task. Performance is both aggregated globally across all tasks (in terms of task-specific metrics) and across subsets of tasks with the same evaluation metric.}
    \label{tab:multitask}
\end{table}

\begin{table*}[p]
    \centering
    \begin{tabularx}{\textwidth}{r|>{\sc}X}
          & eli5-asks, eli5-eli5, ethos-sexual-orientation \\
3 & google-wellformed-query, reddit-tifu-title \\
2 & app-reviews, climate-fever, dbpedia-14, emotion, glue-mrpc, glue-rte, glue-sst2, glue-wnli, hatexplain, imdb, liar, mocha, onestop-english, paws, piqa, poem-sentiment, rotten-tomatoes, scicite, tab-fact, trec-finegrained, tweet-eval-emoji, tweet-eval-sentiment, tweet-eval-stance-abortion, tweet-eval-stance-atheism, tweet-eval-stance-climate, tweet-eval-stance-feminist, wiki-auto, yahoo-answers-topics, yelp-review-full \\
2 3 & ade-corpus-v2-dosage, biomrc, boolq, emo, ethos-disability, hate-speech18, kilt-ay2, lama-conceptnet, lama-google-re, lama-squad, mc-taco, numer-sense, proto-qa, ropes, search-qa, sms-spam, superglue-record, tweet-eval-hate, tweet-eval-irony, tweet-eval-offensive \\
1 & circa, crawl-domain, glue-cola, superglue-rte \\
1 3 & lama-trex, limit, qa-srl, superglue-multirc, tweet-eval-stance-hillary, wikisql \\
1 2 & ai2-arc, anli, aqua-rat, blimp-sentential-negation-npi-licensor-present, codah, ethos-gender, ethos-national-origin, ethos-race, ethos-religion, freebase-qa, glue-mnli, glue-qqp, hellaswag, medical-questions-pairs, openbookqa, quarel, quartz-no-knowledge, quartz-with-knowledge, race-middle, scitail, sick, social-i-qa, superglue-cb, superglue-copa, superglue-wic, superglue-wsc, swag, wiki-qa \\
1 2 3 & adversarialqa, art, commonsense-qa, cos-e, definite-pronoun-resolution, ethos-directed-vs-generalized, hotpot-qa, sciq, squad-with-context, wino-grande, wiqa \\
0 & break-QDMR, break-QDMR-high-level, e2e-nlg-cleaned, eli5-askh, multi-news \\
0 3 & aeslc, common-gen, gigaword, race-high, reddit-tifu-tldr, tweet-qa, wiki-split \\
0 2 & ag-news, kilt-wow \\
0 2 3 & blimp-sentential-negation-npi-scope, discovery, hate-speech-offensive, jeopardy, kilt-hotpotqa, kilt-nq, kilt-trex, kilt-zsre, squad-no-context, web-questions, xsum \\
0 1 & ade-corpus-v2-classification, aslg-pc12, financial-phrasebank, glue-qnli, spider \\
0 1 3 & samsum, trec, wiki-bio \\
0 1 2 & amazon-polarity, blimp-anaphor-gender-agreement, blimp-anaphor-number-agreement, blimp-determiner-noun-agreement-with-adj-irregular-1, blimp-ellipsis-n-bar-1, blimp-ellipsis-n-bar-2, blimp-existential-there-quantifiers-1, blimp-irregular-past-participle-adjectives, blimp-wh-questions-object-gap, cosmos-qa, crows-pairs, dream, kilt-fever, math-qa, quoref \\
0 1 2 3 & acronym-identification, ade-corpus-v2-effect, duorc, empathetic-dialogues, health-fact, qasc, quail, tweet-eval-emotion, yelp-polarity \\

    \end{tabularx}
    \caption{Skill allocation to 120 training CrossFit tasks for $|\calS| = 4$.}
    \label{tab:taskalloc}
\end{table*}

\end{document}